\def\year{2021}\relax
\documentclass[letterpaper]{article} 
\usepackage{aaai21}  
\usepackage{times}  
\usepackage{helvet} 
\usepackage{courier}  
\usepackage[hyphens]{url}  
\usepackage{graphicx} 
\usepackage{graphicx}  
\usepackage{natbib}  
\usepackage{caption} 
\usepackage{amsmath}
\usepackage{amsfonts}
\usepackage{multirow}
\usepackage{caption}
\usepackage{subcaption}
\newcommand{\xxrightarrow}[1]{\mathrel{\raisebox{-3pt}{$\xrightarrow{#1}$}}}
\newif\ifarivx

\usepackage{bm}
\usepackage{xspace}

\DeclareRobustCommand{\uvec}[1]{{%
  \ifcsname uvec#1\endcsname
     \csname uvec#1\endcsname
   \else
    \bm{\hat{\mathbf{#1}}}%
   \fi
}}
\makeatletter
\DeclareRobustCommand\onedot{\futurelet\@let@token\@onedot}
\def\@onedot{\ifx\@let@token.\else.\null\fi\xspace}
 
\def\ie{\emph{i.e}\onedot}

\def\wrt{w.r.t\onedot} 

\makeatother

\title{HopRetriever: Retrieve Hops over Wikipedia to Answer Complex Questions}
\arivxtrue

\author {
        Shaobo Li,\textsuperscript{\rm 1}\thanks{This work was done during an internship at Huawei Noah’s Ark Lab.}
        Xiaoguang Li,\textsuperscript{\rm 2}
        Lifeng Shang,\textsuperscript{\rm 2}
        Xin Jiang,\textsuperscript{\rm 2}\\
        Qun Liu,\textsuperscript{\rm 2}
        Chengjie Sun,\textsuperscript{\rm 1}
        Zhenzhou Ji,\textsuperscript{\rm 1}
        Bingquan Liu\textsuperscript{\rm 1}
        \\
}
\affiliations {
    \textsuperscript{\rm 1}Harbin Institute of Technology \\
    \textsuperscript{\rm 2}Huawei Noah’s Ark Lab   \\
    shli@insun.hit.edu.cn, \{lixiaoguang11, shang.lifeng, Jiang.Xin, qun.liu\}@huawei.com,\\ \{sunchengjie, jizhenzhou, liubq\}@hit.edu.cn
}

\begin{document}
\maketitle

\begin{abstract} 
Collecting supporting evidence from large corpora of text (e.g., Wikipedia) is of great challenge for open-domain Question Answering (QA). Especially, for multi-hop open-domain QA, scattered evidence pieces are required to be gathered together to support the answer extraction. In this paper, we propose a new retrieval target, \textbf{hop}, to collect the hidden reasoning evidence from Wikipedia for complex question answering. Specifically, the hop in this paper is defined as the combination of a hyperlink and the corresponding outbound link document. The hyperlink is encoded as the mention embedding which models the structured knowledge of how the outbound link entity is mentioned in the textual context, and the corresponding outbound link document is encoded as the document embedding representing the unstructured knowledge within it. Accordingly, we build HopRetriever which retrieves hops over Wikipedia to answer complex questions. Experiments on the HotpotQA dataset demonstrate that HopRetriever outperforms previously published evidence retrieval methods by large margins. Moreover, our approach also yields quantifiable interpretations of the evidence collection process. 
\end{abstract}

\section{Introduction}
Multi-hop QA~\cite{hotpot-qa-dataset} is the Question Answering (QA) task that requires reasoning over multiple supporting documents to extract the final answer. For the open-domain setting, a key part of Multi-hop QA is to retrieve an evidence path from the whole knowledge source (e.g., Wikipedia). Most of the recent works view multi-hop evidence collection as an iterative document retrieval problem~\cite{asai2019PR,Multi-Hop-Paragraph-Retrieval,Multi-step-Retriever-Reader}, which can be decomposed to several single-step document retrieval. In contrast, some others~\cite{DrKIT,CogQA} focus on mentioned entities and try to traverse textual data like a virtual structured Knowledge Base (KB). These two methods leverage two different kinds of knowledge for evidence collection respectively: (i) informative but unstructured facts inside the introductory documents of entities. (ii) the structured and implicit relations between entities themselves\footnote{In this paper, we view the entity relation as structured knowledge is because it directly connects two entities and can be applied to build a structured entity graph.}. 


\begin{figure}[t]
    \begin{center}
        \includegraphics[width=0.95\linewidth]{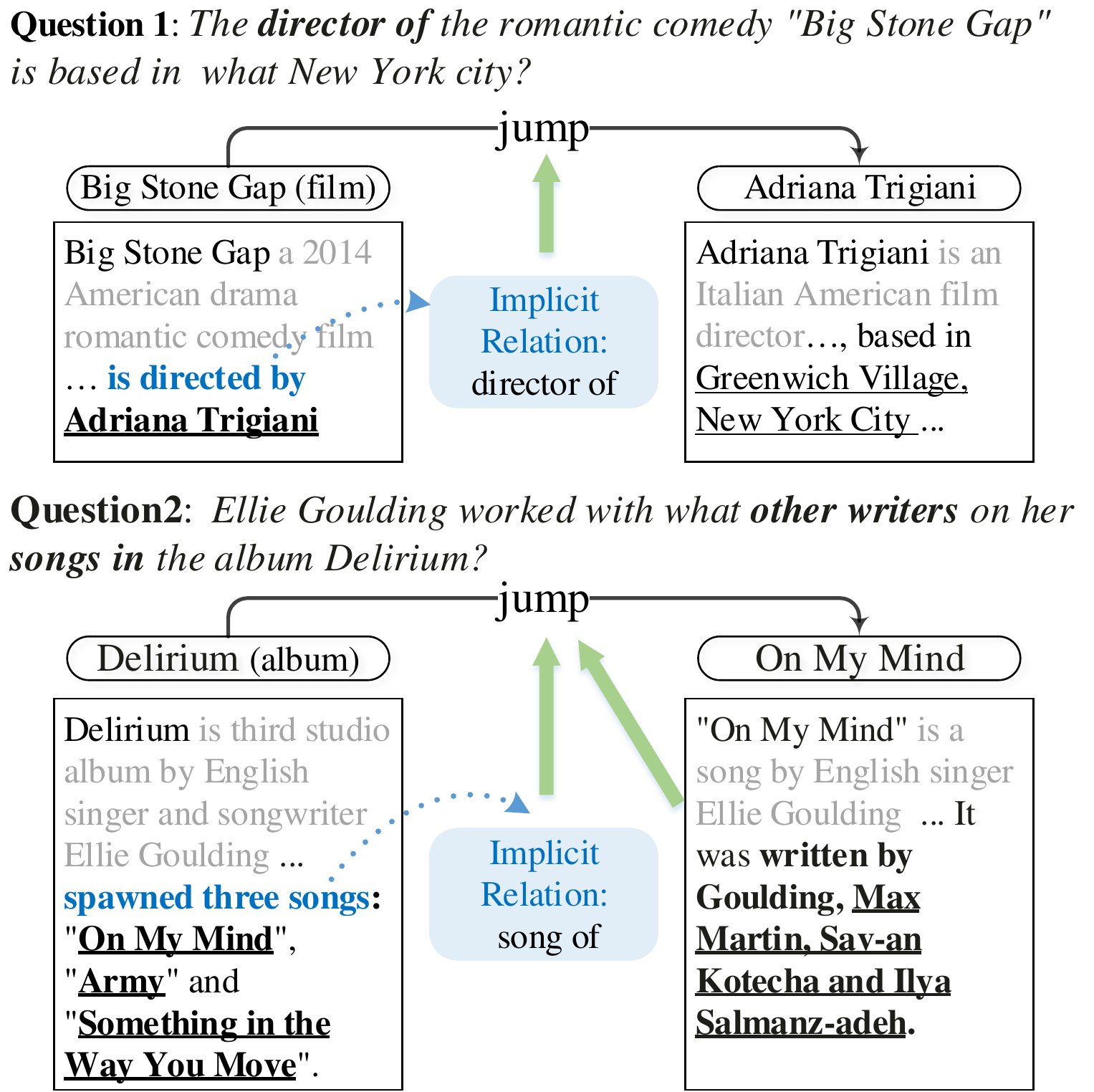}
        \caption {Two examples showing that both structured relation and unstructured fact are needed for complex question answering.}
        \label{fig:intro}
    \end{center}
\end{figure}

Two examples in Figure~\ref{fig:intro} show that both of the above knowledge is needed for complex question answering. We consider the problem of based on what evidence can one jump to the second document for further retrieval. For question 1, the structured relation \textit{``directed by"} implied by \textit{``...directed by Adriana Trigiani"} in the first document matches the relation \textit{``director of''} in the question, hence providing sufficient and convincing evidence that one can hop to the introductory document of \textit{Adriana Trigiani} for further retrieval, even without pre-reading it. However, things become complicated for question 2, for three entities share the same relation \textit{``song of"}: \textit{On My Mind}, \textit{Army}, and \textit{Something in the Way You Move}. In fact, only the entity \textit{On My Mind} satisfies the condition \textit{``works with other writers''} in the question, which makes the relation itself insufficient and indistinctive to make the choice among the three entities. The truth is that only if the unstructured facts about the entity \textit{On My Mind} is browsed through, one can find the conclusive evidence.  

As shown above, to collect sufficient supporting evidence within Wikipedia, it's necessary to consider both relational structures between entities and unstructured knowledge hidden inside the introductory document. When the answering process follows the pattern of \textit{``following the vine to get the melon''}, implicit entity-level relation makes retrieval efficient and effective. However, when the relation chain failed, those unstructured facts in the document mount the stage. 

In this paper, we study how the structured and unstructured knowledge can be combined together and collaboratively contribute to the evidence collection. Accordingly, We define a \textbf{hop} as the combination of a hyperlink and a corresponding outbound link document. A hyperlink in Wikipedia implies how the introductory document of an entity mentions some other, while the outbound link document stores all the unstructured facts and events, which makes a hop contain both relational and factoid evidence for future retrieval. 

One challenge is how to transform the binary (link or not) hyperlink in Wikipedia to distributed representations implying the implicit and complicated entity relation. One step towards this is the recent work on distributed relation learning \cite{matching-the-blanks}, in which the relation representations are learned solely from the entity-linked text in an unsupervised way. With the powerful ability of BERT~\cite{BERT} for text encoding, \citep{CogQA} and \cite{DrKIT} encodes entity spans into node representations to conduct relation-based reasoning. In this paper, we represent each hyperlink with the corresponding entity mention, with the currently described entity as the mention subject and the outbound link entity as the mention object.

\textbf{Our contributions.} To be more specific, this paper introduces HopRetriever, a method to automatically and adaptively leverage both the structured entity relation and unstructured introductory facts for evidence collection. For each entity mention within Wikipedia documents, we encode the textual context around it into mention embedding to represent the implicit structured knowledge. As for the representation of unstructured knowledge in documents, we use BERT to encode document text conditioned on the original question, as previous works do. For each step retrieval, the hop from one document(entity) to another one can gather evidence from two perspectives: (i) How the current document mentions the other one. (ii) What facts are hidden in the introductory document of the other entity. Experiments conclude that our retrieval method outperforms both entity-centric retrieval methods and document-wise retrieval ones.

Our prime contributions are as follows: 
\begin{itemize}
\item We propose to retrieve hops over Wikipedia to answer complex questions, which adaptively and selectively collects evidence from both structured entity relation and unstructured facts within documents. 

\item We propose to represent hyperlinks in Wikipedia with mention embeddings, which we show can precisely capture the implicit relation between entities. 

\item Evaluated on HotpotQA~\cite{hotpot-qa-dataset}, the proposed approach significantly outperforms previously published evidence retrieval methods. Additionally, we conduct further experimental analysis and demonstrate the good interpretability of our method.

\end{itemize}

\section{Related Works}
\label{sec:related-works}
\textbf{Document-wise reasoning.}
Most current open-domain QA methods directly retrieve documents for evidence collection. \citet{DRQA}, \citet{ORQA}, \citet{DPR}, and \citet{SemanticMRS} leverage sparse methods like BM25 or fully trainable models to retrieve candidates from the whole Wikipedia collection. Such approaches, however, find evidence documents independently without knowing the previous retrieved ones, which may cause retrieval failures when one of the evidence documents has a little semantic relationship with the original question. Avoiding this, \citet{Multi-Hop-Paragraph-Retrieval} and \citet{Multi-step-Retriever-Reader} introduce multi-step retrievers to explore multiple evidence documents iteratively. Most recently, \citet{asai2019PR} proposes the PathRetriever that retrieves documents paths along the outbound-link of text graph. With the graph structure of the documents, PathRetriever reduces the search space of documents during each step retrieval, which is much smaller than that of previous iterative retrievers. The biggest difference between PathRetriever and our method is that we additionally consider the structured and multi-valued relation between entities, while PathRetriever uses hyperlinks in a binary way: link or not link.

\begin{figure*}[h]
    \centering
    \includegraphics[width=0.95\linewidth]{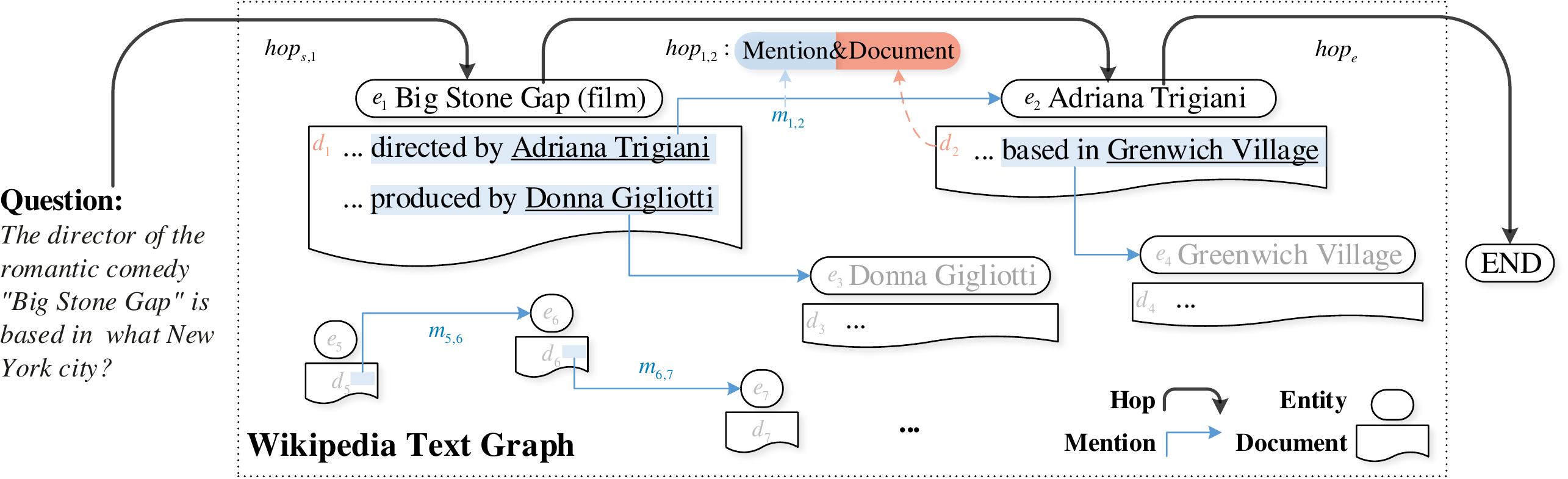}
    \caption{Retrieving Hops over Wikipedia text graph. Documents are retrieved by selecting hops over them iteratively. Each directed arrow implies a mention $m_{i,j}$, which reveals how $e_{i}$ mentions $e_{j}$ in the document $d_{i}$. Hops between entities are indicated by curved arrows. If the mention $m_{i,j}$ exists between $e_{i}$ and $e_{j}$, the hop $hop_{i,j}$ is represented based on both $m_{i,j}$ and the introductory document $d_j$ for retrieval or based on the $d_j$ solely if no mentions exist. 
    } 
    \label{fig:overview}
\end{figure*}

\noindent
\textbf{Entity-centric reasoning.}
Considering that most factoid QA problems are entity-centric, some other works focus on the entity mention to collect reasoning evidence. Cognitive Graph~\cite{CogQA} trains a reading comprehension model to predict the next-hop spans, aiming to find the most evidential mentioned entity. Similarly, DrKIT~\cite{DrKIT} constructs large mounts of entity mentions from the corpus and proposes a method to reason over these mentions, softly following paths of latent relations. We've shown in Figure~\ref{fig:intro} that when the question is not the case of \textit{``following the vine to get the melon"}, the mention itself fails to provide sufficient reasoning evidence for which entity to hop. Inspired by the idea of pseudo-relevance feedback~\cite{prf}, \citet{Multi-step-Entity-centric} also leverages entity-link to find more supporting evidence. However, this method is still document-level, for the entity links are used not for relation representation, but document expansion. We empirically show significant improvement over the above methods.

\noindent
\textbf{Question decomposition.}
\citet{QDMR}, \citet{Unsupervised-QD}, and \citet{DecompRC} propose to decompose a complicated question into several simpler sub-questions and conduct single-hop QA at each step. The challenge for question decomposition is to ensure each sub-question collects the truly necessary evidence. As we know from example 2 in Figure~\ref{fig:intro}, when the structured relation fails, one can not ask reasonable sub-question without exploring enough introductory documents at the next step.


\section{Method}
\subsection{Overview}
\subsubsection{Task definition.}
Our task is to obtain the answer $a$ for an open-domain multi-hop question $q$. A retriever model $\rm Retriever$ is used to collect the multiple evidence pieces over a large-scale knowledge source $K$:
\begin{equation}
    D_q={\rm Retriever}(q,K).
\end{equation}
$D_q$ should contain multiple documents that are necessary for answering the multi-hop question. All textual facts in $D_q$ and $q$ are concatenated together and fed into a answer extraction model $\rm Reader$ to obtain the answer $a$:
\begin{equation}
    a={\rm Reader}(q,D_q).
\end{equation}

\subsubsection{Our approach.}
In this paper, we propose HopRetriever to take the place of the retriever model $\rm Retriever$ while keeping the answer extraction model $\rm Read$ as standard \cite{BERT}. The knowledge source $K$ is constructed from Wikipedia\footnote{\url{https://en.wikipedia.org}}. Each Wikipedia page corresponds to an entity $e_i$, accompanied by an introductory document $d_i$. Moreover, if there exists an anchor text in $d_i$ linked to $e_j$, we denote it as a mention $m_{i,j}=e_i\xxrightarrow{d_i} e_j$, which means $e_j$ is mentioned by $e_i$ via $d_i$. Accordingly, the knowledge source is formulated as $K=\{D,E,M\}$ that consists of an entity set $E=\{e_i\}$, an introductory document set $D=\{d_i\}$, and a mention set $M=\{m_{i,j}\}$. 

$D_q$ is retrieved iteratively. At each retrieval step, a document is fetched by examining not only the unstructured facts contained in but also the mention of it in the latest selected document. To achieve that, we encode the unstructured textual facts and the mention respectively and then represented them together within a hop. HopRetriever uses hops as the matching objects when retrieving over Wikipedia. The overview of a retrieval process is shown in Figure~\ref{fig:overview}. The details about the hop encoding and the iterative retrieval procedure are discussed in the following two sections.

\subsection{Hop Encoding}
\label{sec:hop_enc}

HopRetriever considers retrieving a new document $d_j$ conditioning on the retrieval history as finding the proper hop from the current foothold entity $e_i$ to the entity $e_j$. The representation of each hop consists of mention embedding $\mathbf{m}_{i,j}$ that implies the entity relation from $e_{i}$ to $e_j$, and the document embedding $\mathbf{u}_j$ of the introductory document of entity $e_j$. 

\subsubsection{Mention embedding.}
\begin{figure}[h]
    \centering
    \includegraphics[width=\columnwidth]{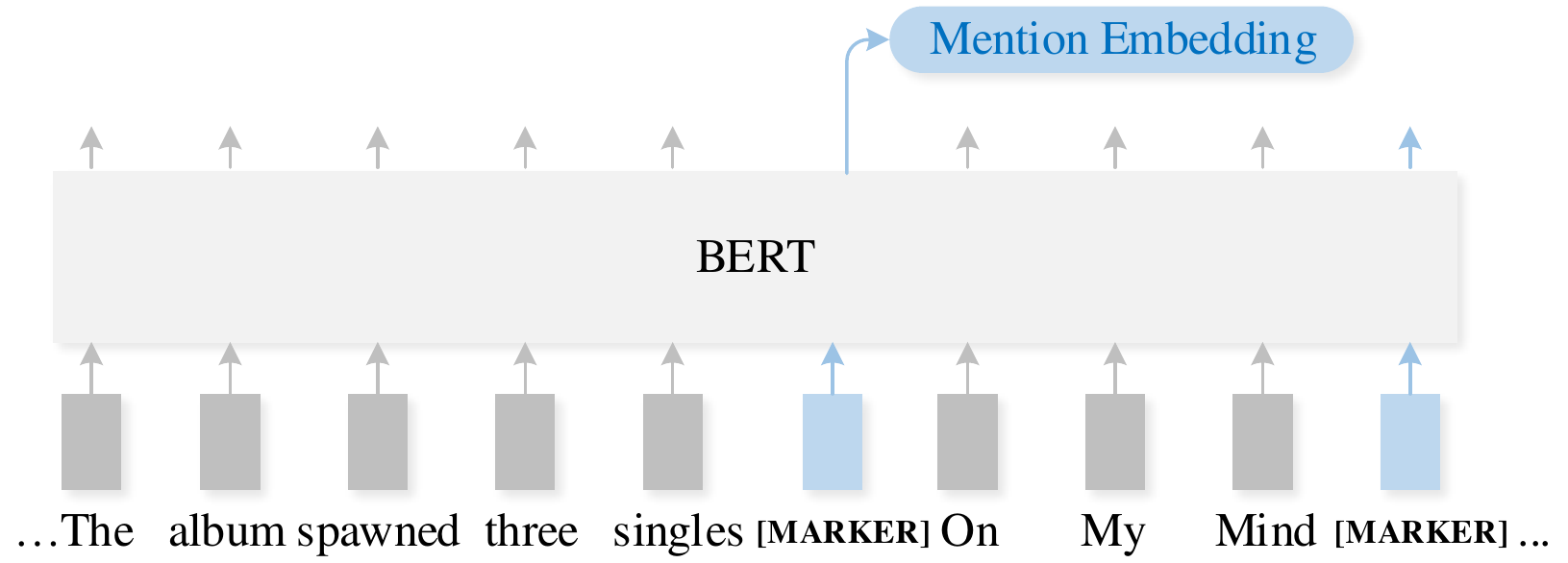}
    \caption{Encoding the mention using entity markers.}
    \label{fig:entity_marker}
\end{figure}

We consider the problem of how to encode a hop $hop_{i,j}$ into hop encoding $\mathbf{hop}_{i,j}$. The structured entity relation revealed by $m_{i,j}$ is encoded as mention embedding $\mathbf{m}_{i,j}$, based on the context around it. Inspired by \citet{matching-the-blanks}, two entity markers clipping the anchor text of each mentioned entity are introduced to obtain the mention embedding. An example is shown in Figure~\ref{fig:entity_marker} (from the second example in Figure~\ref{fig:intro}), the document that contains the mention of \textit{On My Mind} is fed into BERT with two additional \textsc{[marker]} tokens, and the output representation of the first \textsc{[marker]} token is used as the mention embedding vector.

\noindent If $e_j$ is not mentioned directly in the introductory document of $e_{i}$, we represent the relation between them with a trainable uniformed vector $\mathbf{m}_\text{\sc p}$, as shown below:
\begin{equation}
    \mathbf{m}_{i,j} = 
    \begin{cases}
        {\rm BERT}_\text{\sc [m-$j$]}(q;d_{i}),   & \text{if $m_{i,j}\in M$} \\
        \mathbf{m}_\text{\sc p},                  & \text{otherwise} 
    \end{cases}
    \label{eq:mention_emb}
\end{equation}
where the ${\rm BERT}_\text{\sc [m-$j$]}$ is the representation of the entity marker corresponding to entity $e_j$.

\subsubsection{Document embedding.}
The unstructured knowledge about the entity $e_j$ is encoded as document embedding $\mathbf{u}_j$ by feeding the textual facts in $d_j$ (concatenated with $q$) into BERT, and the output representation of the \textsc{[cls]} token is taken as the document embedding vector: 
\begin{equation}
\mathbf{u}_j = {\rm BERT}_\text{\sc [cls]}(q;d_j).
\end{equation}

\subsubsection{Knowledge fusion.}
The mention embedding $\mathbf{m}_{i,j}$ and the document embedding $\mathbf{u}_j$ are fused together as hop encoding $\mathbf{hop}_{i,j}$ by the attention mechanism proposed in \citet{sukhbaatar2015end}. The following fusion procedure allows HopRetriever to adaptively and selectively manage the two kinds of knowledge according to which truly matters: 
\begin{equation*}
    \begin{gathered}
        a_m                = \mathbf{h}\mathbf{W}_k \mathbf{m}_{i,j}    \\
        a_u                = \mathbf{h}\mathbf{W}_k \mathbf{u}_{j}        \\
        \{w_m,w_u\}        = {\rm softmax}(\{a_m,a_u\})                       
    \end{gathered}
\end{equation*}
\begin{equation}
        \mathbf{hop}_{i,j} = w_m\cdot \mathbf{W}_v \mathbf{m}_{i,j}  + w_u \cdot \mathbf{W}_v \mathbf{u}_{j} ,
    \label{eq:fusion}
\end{equation}
where $\mathbf{h}$ is the vector that encodes the corresponding retrieval history, the $\mathbf{W}_k$ projects the two embedding vectors (\ie $\mathbf{m}_{i,j}$ and $\mathbf{u}_{j}$) into \textit{key} vectors. The $\mathbf{h}$ acts as \textit{query} vector that interacts with the \textit{key} vectors to calculate the importance weight $w_m$ for the mention embedding $\mathbf{m}_{i,j}$ and $w_u$ for the document embedding $\mathbf{u}_j$, then $\mathbf{m}_{i,j}$ and $\mathbf{u}_j$ are projected into \textit{value} vectors by $\mathbf{W}_v$ and fused as hop encoding with important weights.

\subsection{Iterative Retrieval of Hops}

\begin{figure}[h]
    \centering
    \includegraphics[width=\columnwidth]{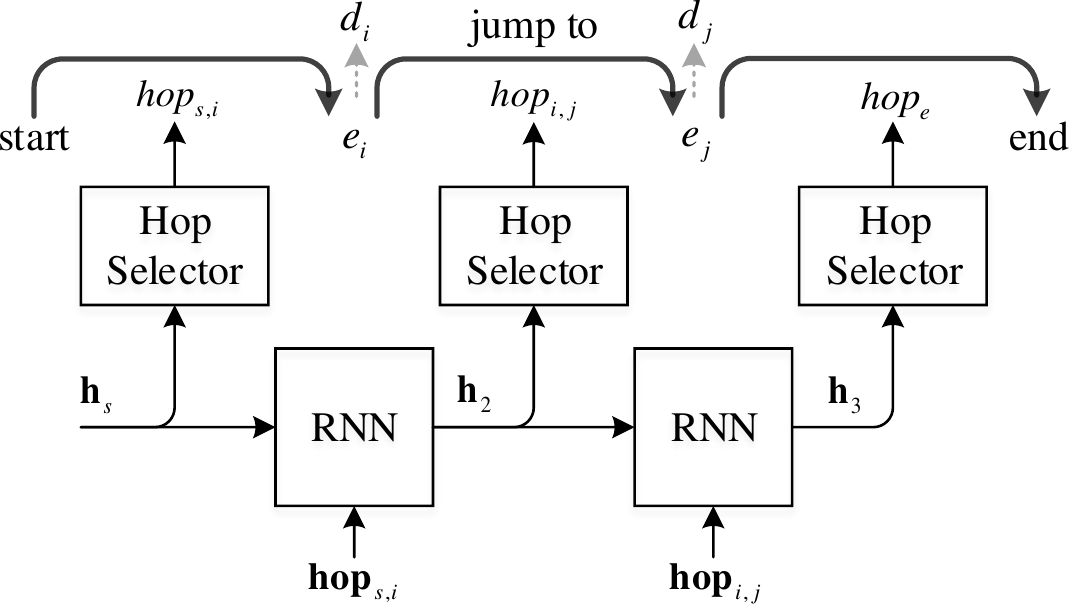}
    \caption{The retrieval process of HopRetriever for three hops. $hop_{s,i}$ indicates a beginning jump from the start to $e_i$ is selected based on the initial hidden state $\mathbf{h}_{s}$. The selection of hop $hop_{i,j}$ retrieves the supporting document $d_j$ at the second step. $hop_{e}$ ends the retrieval process finally.}
    \label{fig:rnn}
\end{figure}
Figure~\ref{fig:rnn} illustrates a three-step recurrent hop retrieval process. Generally, let $e_i$ denote the foothold entity selected at the previous $t-1$ step, the probability of retrieving the document $d_j$ at $t$ step is calculated by the dot product of $\mathbf{h}_t$ and hop encoding $\mathbf{hop}_{i,j}$ (\ie the Hop Selector in Figure~\ref{fig:rnn}),- as formulated in the following equation:
\begin{equation}
    \begin{gathered}
        p(d_j) = {\rm sigmoid}(\mathbf{h}_t^\top\mathbf{hop}_{i,j}),
    \end{gathered}
\end{equation}
\noindent
where $\mathbf{h}_t$ is the hidden state vector that encodes all the previously selected hops by a Recurrent Neural Network (RNN): 
\begin{equation}
    \mathbf{h}_t =
    \begin{cases}
         \mathbf{h}_s,                                                     & t = 1 \\
         {\rm RNN} (\mathbf{h}_{t-1},\mathbf{hop}_{k,i}),                  & t \ge 2 
    \end{cases}
\end{equation}
where $\mathbf{h}_s$ is the initial hidden state vector and $\mathbf{hop}_{k,i}$ is the encoding of the hop selected at $t-1$ step. Specially, for $t=1$, the hop $hop_{s,j}$ indicating jumping from the retrieving start to $e_j$ is introduced. Similarly, a special end hop ${hop}_{e}$ is used to mark the end of the retrieval process and it is encoded by $\mathbf{m}_p$ and a virtual end document encoding $\mathbf{u}_e$. Let $f$ denote the fusion function formulated as Equation~\eqref{eq:fusion}, the encodings of different hops are summarized in Table~\ref{tab:hop_type}.

\begin{table}[ht]
    \footnotesize
    \centering
    \begin{center}
    \begin{tabular}{c|c|c}
    \hline 
    \textbf{Notation}                & \textbf{Encoding}                         & \textbf{Explanation}              \\ \hline
    \multirow{2}{*}{${hop}_{i,j}$}   & $f(\mathbf{m}_\text{\sc p},\mathbf{u}_j)$ & $e_j$ is not mentioned in $d_{i}$ \\
                                     & $f(\mathbf{m}_{i,j},\mathbf{u}_j)$        & $e_j$ is mentioned in $d_{i}$       \\\hline
    ${hop}_{s,j}$                    & $f(\mathbf{m}_\text{\sc p},\mathbf{u}_j)$ & Select $d_j$ at the beginning     \\
    ${hop}_{e}$                      & $f(\mathbf{m}_\text{\sc p},\mathbf{u}_e)$ & Retrieval finish                  \\ \hline
    \end{tabular}
    \end{center}
    \caption{Types of hop encoding.}
    \label{tab:hop_type}
\end{table}

\subsection{Fine-Grained Sentence-Level Retrieval}
A single supporting document can be split into multiple sentences and may not all these sentences are essential for answering the question. Pointing out the indispensable supporting sentences can illuminate the reasons why a document is required. In HopRetriever, the supporting sentence prediction is added as an auxiliary task along with the primary hop retrieval task. At step $t$, the probability $p(s_{i,l})$ that indicates the $l$-th sentence in the latest retrieved document $d_{i}$ is a supporting sentence is calculated by the following equations:
\begin{equation}
        \mathbf{s}_{i,l} = {\rm BERT}_\text{{\sc [sm-$l$]}}(q;d_{i})
\end{equation}
\begin{equation}
        p(s_{i,l}) ={\rm sigmoid} (\mathbf{h}_t\mathbf{W}_s\mathbf{s}_{i,l}) ,
\end{equation}
where $\mathbf{s}_{i,l}$ is the sentence embedding vector obtained by inserting a sentence marker \textsc{[sm-$l$]} at the end of the $l$-th sentence in $d_{i}$, which is similar to how the mention embedding is obtained. If $p(s_{i,l}) > 0.5$, then the $l$-th sentence in document $d_{i}$ is identified as a supporting sentence.

\subsection{Objective Functions of HopRetriever}
HopRetriever is a sequence prediction model with binary cross-entropy objective functions at each step. At the retrieval step $t$, the objective function of the primary hop retrieval task is
\begin{equation}
    \label{eq:obj_doc}
    \log p(d_j)+\sum_{\bar d_j\in D, \bar d_j \ne d_j }{\log(1-p(\bar d_j))} ,
\end{equation}
where $d_j$ is the ground-truth document. For the auxiliary supporting sentence prediction task, the object function at step $t$ is
\begin{equation}
    \label{eq:obj_sent}
    \sum_{l \in L_i}{\log p(s_{i,l})}+\sum_{l \notin L_i}{\log(1-p(s_{i,l}))} ,
\end{equation}
where $s_{i,l}$ is the $l$-th sentence in $d_i$, $L_i$ is the set of indices of the ground-truth supporting sentences in $d_i$. The above two objective functions are maximized together in training.


\begin{table*}[htbp]
    \footnotesize
    \centering
    \begin{tabular}{l|ccc|ccc|ccc}
        \hline
        \multicolumn{1}{l|}{\textbf{Model}}                        &\multicolumn{3}{c|}{\textbf{Ans exists}}     & \multicolumn{3}{c|}{\textbf{Sent exists}}   & \multicolumn{3}{c}{\textbf{All docs exist}} \\ 
        \multicolumn{1}{r|}{}                                      & top-1        & top-5        & top-8         & top-1        & top-5        & top-8         & top-1        & top-5      & top-8           \\ 
        \hline 
        Cognitive Graph QA \cite{CogQA}                            &72.21         &-             &-              &70.25         &-             &-              &57.80         &-           &-                \\
        Semantic Retrieval\textsuperscript{*} \cite{SemanticMRS}   &77.84         &85.96         &86.39          &81.68         &88.35         &88.50          &69.35         &81.73       &82.07            \\
        PathRetriever \cite{asai2019PR}                            &80.96         &89.09         &89.98          &82.05         &88.69         &89.48          &73.91         &86.12       &87.39            \\
        \hline
        \textbf{HopRetriever}                                      &\textbf{86.89}&\textbf{91.11}&\textbf{91.80} &\textbf{88.41}&\textbf{92.78}&\textbf{93.20} & \textbf{82.54}&\textbf{88.60}&\textbf{89.09}\\
        \ifarivx
        HopRetriever-plus                                          &89.25         &93.07         &93.64          &91.29         &95.44         &95.70          &86.94         &93.25       &93.72            \\
        \fi
        \hline
    \end{tabular}
    \caption{Evidence collection result on the HotpotQA fullwiki development set. We compare the top-1, top-5, and top-8 output document sequences from different retrievers using respectively. Cognitive Graph QA produces one document sequence for each question. HopRetriever and PathRetriever output the top-8 document sequences by the adoption of beam search. Semantic Retrieval\textsuperscript{*} ranks documents instead of document sequences, so we assemble the top-2, top-10, and top-16 output documents into the top-1, top-5, and top-8 document sequences respectively for better fairness.}
    \label{tab:doc_res}
\end{table*}

\section{Experiments}
\subsection{Setup}
\subsubsection{Dataset.} 
HopRetriever is evaluated on the multi-hop question answering dataset HotpotQA \cite{hotpot-qa-dataset}, which includes 90,564 question-answer pairs with annotated supporting documents and sentences for training, 7,405 question-answer pairs for development, and 7,405 questions for testing. All the testing questions correspond to multiple supporting documents. We focus on the \textit{fullwiki} setting of HotpotQA, in which the supporting documents for each question are scattered among almost 5M Wikipedia pages. In the official evaluation, the participant model is required to predict both the exact supporting sentences and the answer text.

\subsubsection{Pipeline.}
The whole procedure follows a coarse-to-fine pipeline that contains three stages: 
\begin{enumerate}
    \item Preliminary retrieval: Only the top-500 documents are used to construct the initial candidate hops of HopRetriever, according to the TF-IDF scores of documents \wrt the input question.
    \item Supporting documents retrieval and supporting sentence prediction: HopRetriever retrieves the supporting documents iteratively starting from the initial candidate hops, and also predicts supporting sentences from the retrieved documents.  
    \item Answer extraction: The answer within the retrieved supporting documents is extracted using BERT (large, whole word mask), following the conventional answer boundary prediction approach~\cite{BERT,seobi}, which is the same as PathRetriever \cite{asai2019PR}.
\end{enumerate}

\subsubsection{Implementation details.}
The negative hop sequences used to train the proposed model are constructed by traversing through the entities in Wikipedia. And the top-40 TD-IDF scored documents \wrt the question and top-10 scored documents \wrt the ground-truth documents are used as the start points of the traverse.The length of negative hop sequences is fixed to 3. We restrict the maximum input sequence length of BERT to 384. In training, the batch size is set to 16, the learning rate is $3\times 10^{-5}$, and the number of training epochs is 3. We use beam search with beam size set to 8 at the inference time. 

\ifarivx
To achieve better performance, we introduce a neural ranker based on BERT-base \cite{BERT4ReRank} to produce more precise top-500 documents in the preliminary retrieval. And use ELECTRA \cite{ELECTRA} to take the place of BERT, i.e., use the ELECTRA base in HopRetriever for document sequence retrieval and use ELECTRA large for answer extraction. The results of this enhanced pipeline are denoted as HopRetriever-plus.
\fi

\begin{table*}[htbp]
    \footnotesize
    \centering
    \begin{tabular}{l|ccc|ccc|ccc}
        \hline
        \multicolumn{1}{l|}{\textbf{Model}}                        &\multicolumn{3}{c|}{\textbf{Ans exists}}     & \multicolumn{3}{c|}{\textbf{Sent exists}}   & \multicolumn{3}{c}{\textbf{All docs exist}}\\ 
        \multicolumn{1}{r|}{Recall @}                              & top-1         & top-5         & top-8         & top-1         & top-5        & top-8         & top-1        & top-5         & top-8         \\ \hline 
        PathRetriever (Comparison)                                 &77.00          &\textbf{81.17} &82.25          &88.33          &90.36         &90.62          &\textbf{86.42}&\textbf{89.58} &\textbf{90.38} \\
        \textbf{HopRetriever} (Comparison)                         &\textbf{77.40} &80.97          &\textbf{82.31} &\textbf{91.31} &\textbf{92.73}&\textbf{92.79} &84.26         &85.41          &85.41          \\ \hline
        PathRetriever (Bridging)                                   &81.95          &91.08          &91.92          &80.56          &88.29         &89.20          &70.77         &85.25          &86.63          \\ 
        \textbf{HopRetriever} (Bridging)                           &\textbf{89.27} &\textbf{93.66} &\textbf{94.19} &\textbf{87.73}&\textbf{92.79}&\textbf{93.29} &\textbf{82.11}&\textbf{89.41} &\textbf{90.01} \\
        \hline
    \end{tabular}
    \caption{Evidence collection results on different types of questions.}
    \label{tab:doc_res_qt}
\end{table*}

\begin{table*}[htbp]
    \centering
    \footnotesize
    \begin{tabular}{l|l|cc|cc|cc}
        \hline
                                & \multirow{2}{*}{\textbf{Model}}              & \multicolumn{2}{c|}{\textbf{Ans}} & \multicolumn{2}{c|}{\textbf{Sup}} & \multicolumn{2}{c}{\textbf{Joint}} \\  
                                &                                              & EM              & F1              & EM              & F1              & EM               & F1              \\ \hline
        \multirow{4}{*}{dev}  & Cognitive Graph QA \cite{CogQA}                & 37.55           & 49.40           & 23.11           & 58.52           & 12.18            & 35.28           \\ 
                                & Semantic Retrieval \cite{SemanticMRS}        & 46.41           & 58.70           & 39.86           & 71.53           & 26.53            & 49.00           \\ 
                                & PathRetriever \cite{asai2019PR}              & 60.49           & 73.30           & 49.16           & 76.05           & 35.82            & 61.43           \\ \cline{2-8} 
                                & \textbf{HopRetriever}                        & \textbf{62.07}  & \textbf{75.18}  & \textbf{52.53}  & \textbf{78.92}  & \textbf{37.81}   & \textbf{64.50}  \\ 
                                \ifarivx
                                & HopRetriever-plus                            & 66.56           & 79.21           & 56.02           & 81.81           & 42.01            & 68.97           \\ 
                                \fi
        \hline
        \multirow{9}{*}{test}   & DecompRC \cite{DecompRC}                     & 30.00           & 40.65           & -               & -               & -                & -               \\ 
                                & Cognitive Graph QA \cite{CogQA}              & 37.12           & 48.87           & 22.82           & 57.69           & 12.42            & 34.92           \\ 
                                & DrKIT \cite{DrKIT}                           & 42.13           & 51.72           & 37.05           & 59.84           & 24.69            & 42.8            \\ 
                                & Semantic Retrieval \cite{SemanticMRS}        & 45.32           & 57.34           & 38.67           & 70.83           & 25.14            & 47.60           \\ 
                                & Transformer-XH \cite{Trans-XH}               & 51.60           & 64.07           & 40.91           & 71.42           & 26.14            & 51.29           \\ 
                                & PathRetriever \cite{asai2019PR}              & 60.04           & 72.96           & 49.08           & 76.41           & 35.35            & 61.18           \\ 
                                & Semantic Retrieval + HGN \cite{HGN}          & 59.74           & 71.41           & 51.03           & 77.37           & 37.92            & 62.26           \\ \cline{2-8} 
                                & \textbf{HopRetriever}                        & \textbf{60.83}  & \textbf{73.93}  & \textbf{53.07}  & \textbf{79.26}  & \textbf{38.00}   & \textbf{63.91}  \\ 
                                \ifarivx
                                & HopRetriever-plus                            & 64.83           & 77.81           & 56.08           & 81.79           & 40.95            & 67.75           \\ 
                                \fi
        \hline
    \end{tabular}
    \caption{Answer extraction and supporting sentence prediction result in the fullwiki setting of HotpotQA.}
    \label{tab:sa_result}
\end{table*}

\subsection{Results}
\subsubsection{Evidence collection.}
The HopRetriever is first evaluated by measuring the coverage of ground-truth answers, supporting sentences, and supporting documents in the retrieved supporting documents, as shown in Table~\ref{tab:doc_res}. The metric Ans exists measures the percentage of the questions whose answers are extractable from the retrieved document sequence. Sent exists is the percentage of the supporting sentences that can be found. The percentage of the questions that have all ground-truth documents retrieved are showed as the All docs exist. 

Three models that mainly focus on evidence collection over Wikipedia are evaluated as baselines on the development set:
\begin{itemize}
    \item \textbf{Cognitive Graph QA} \cite{CogQA} explicitly utilizes the structured knowledge in Wikipedia with a graph whose nodes are entities or answer spans. The representations of nodes are maintained by Graph Neural Network (GNN) \cite{GNN1,GNN2}. 
    \item \textbf{Semantic Retrieval} \cite{SemanticMRS} is a multi-grained retrieval baseline that retrieves supporting documents and sentences together, focuses on the unstructured knowledge in Wikipedia. 
    \item \textbf{PathRetriever} \cite{asai2019PR} introduces a similar iterative retrieval framework, but only focuses on the unstructured knowledge provided in the introductory document at each retrieval step. 
\end{itemize}

To be fairly compared with PathRetriever, which is the state-of-the-art published model, HopRetriever uses the same initial search space (\ie top-500 documents based on TF-IDF scores) and pre-trained model (\ie BERT-base) with  PathRetriever. Notably, HopRetriever outperforms the PathRetriever by $5.93\%$, $6.36\%$, and $8.63\%$ on the top-1 evidence collection metrics respectively, and also achieves significant improvement over Semantic Retriever and Cognitive Graph QA, which further demonstrates the effectiveness of HopRetriever.

A more detailed comparison with PathRetriever is shown in Table~\ref{tab:doc_res_qt}. We can observe that HopRetrieve works more effectively on the bridging questions. In the HotpotQA dataset, the ground-truth supporting documents of comparison questions may not be directly relevant to each other where no structured knowledge is available, which makes HopRetriever perform almost the same as PathRetriever. In contrast, the ground-truth supporting documents of the bridging questions are stringed with mentions that can provide informative structured knowledge, so HopRetriever performs better by leveraging mentions additionally.

\subsubsection{Answer extraction and supporting sentence prediction.}
Table~\ref{tab:sa_result} shows the performance of different methods for the answer and supporting sentence prediction. Naturally, the answer extraction and supporting sentence prediction result benefit from the improvements of document retrieving. By providing more accurate supporting documents, HopRetriever outperforms all the aforementioned baseline models on the development set and also the other published models\footnote{By the submission time of this paper, recently published method on HotpotQA fullwiki leaderboard is PathRetriever.} on the test set.

\subsection{Analysis}
Detailed analysis of HopRetriever is carried out in this section, especially about how the structured and unstructured knowledge in Wikipedia contribute to evidence retrieval.

\subsubsection{Embedding weights on different question types.}
At the $t$ step in the retrieval procedure of HopRetriever, the decision whether to select a document $d_j$ depends on the hop encoding $\mathbf{hop}_{i,j}$, which contains a mention embedding and a document embedding assigned with learnable weights as formulated in Equation~\eqref{eq:fusion}. We analyze the weights and find that they provide intuitive explanation about which embedding is more important for different question types. Table~\ref{tab:hop_weights} shows the average weight of mention embedding and document embedding on different question types.

\begin{table}[htbp]
    \centering
    \footnotesize
    \begin{tabular}{c|c|c}
        \hline
        \textbf{Question Type}        & \textbf{Mention}    & \textbf{Document}  \\ \hline
        Bridging                      & 89.53\%             & 10.47\%            \\
        Comparison                    & 4.61\%              & 95.39\%            \\
        \hline
    \end{tabular}
    \caption{Weights of mention embedding and document embedding on bridging questions and comparison questions.} 
    \label{tab:hop_weights}
\end{table}

It can be seen that the mention embedding accounts for a large portion (89.53\%) on the bridge questions. The bridge questions always require selecting a proper document along with a hyperlink and the mentions do provide helpful information for bridging the evidence pieces. Conversely, when processing the comparison questions, the weight of mention embedding is relatively small (4.61\%) because there are no available mentions between the supporting documents. 

\begin{figure*}[h]
    \centering
    \begin{subfigure}[b]{0.3\textwidth}
        \centering
        \includegraphics[width=\textwidth]{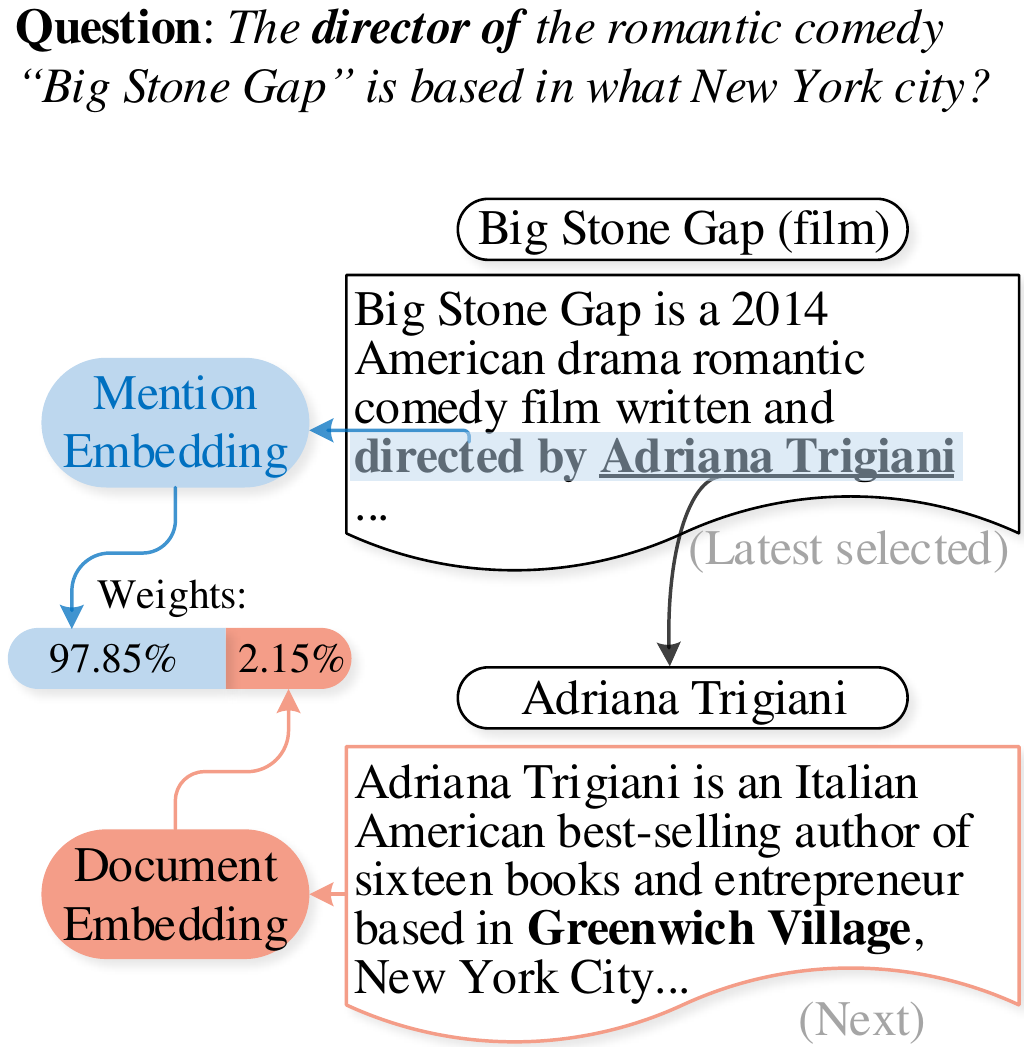}
        \caption{Case 1.}
        \label{fig:case_1}
    \end{subfigure}
    \begin{subfigure}[b]{0.3\textwidth}
        \centering
        \includegraphics[width=\textwidth]{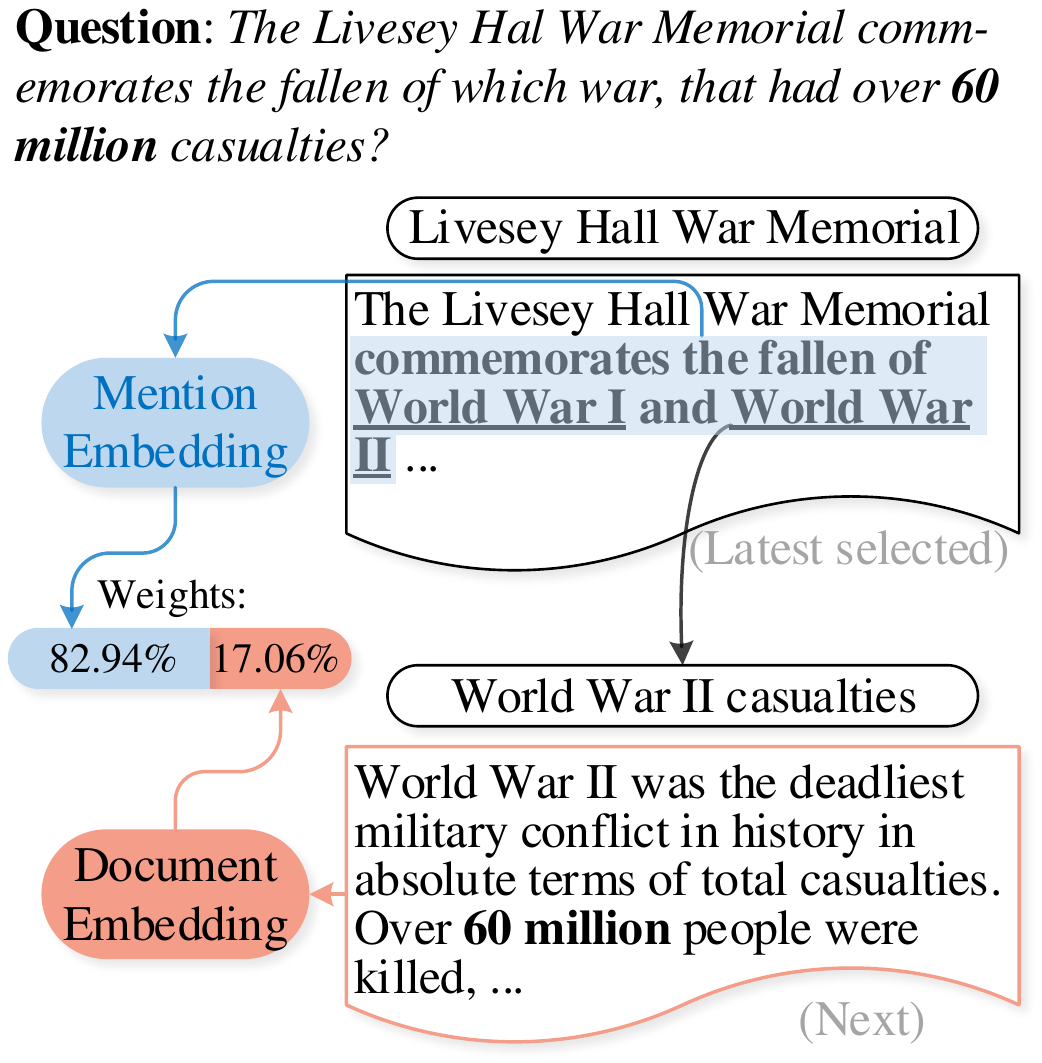}
        \caption{Case 2.}
        \label{fig:case_2}
    \end{subfigure}
    \begin{subfigure}[b]{0.3\textwidth}
        \centering
        \includegraphics[width=\textwidth]{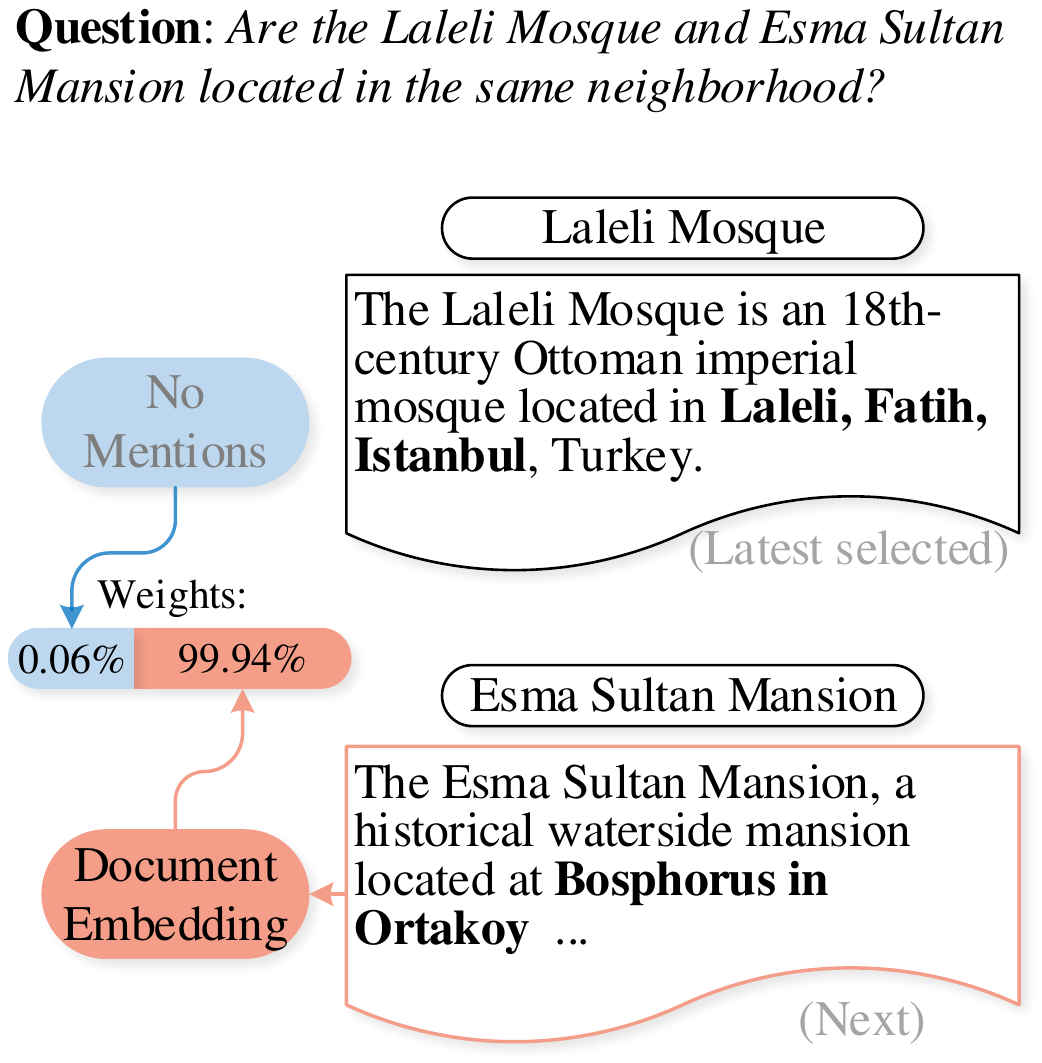}
        \caption{Case 3.}
        \label{fig:case_3}
    \end{subfigure}

\caption {The weights of mention embedding and document embedding in different cases.}
\label{fig:cases}    
\end{figure*}

\begin{table*}[ht]
    \footnotesize
    \centering
    \begin{tabular}{l|ccc|ccc|ccc}
        \hline
        \multicolumn{1}{l|}{\textbf{Model}}         & \multicolumn{3}{c|}{\textbf{Ans exists}} & \multicolumn{3}{c|}{\textbf{Sent exists}} & \multicolumn{3}{c}{\textbf{All docs exist}}   \\ 
        \multicolumn{1}{r|}{Recall @}               & top-1        & top-5       & top-8       & top-1        & top-5       & top-8       & top-1        & top-5       & top-8              \\ \hline
        full                                        & 86.89        & 91.11       & 91.80       & 88.41        & 92.78       & 93.20       & 82.54        & 88.60       & 89.09              \\ \hline
        1. w/o structured knowledge                 & 76.35        & 86.02       & 88.12       & 80.91        & 88.49       & 89.92       & 66.20        & 78.89       & 81.23              \\
        2. w/o weighting                            & 86.21        & 91.07       & 91.52       & 87.73        & 92.55       & 93.09       & 81.38        & 88.09       & 88.70              \\
        3. w/o sentence prediction                  & 86.58        & 90.88       & 91.51       & 87.98        & 92.54       & 92.98       & 82.03        & 88.29       & 88.89               \\
        \hline
    \end{tabular}
    \caption{Ablation experiments of HopRetriever.}
    \label{tab:abl_res}
\end{table*}

\subsubsection{Embedding weights in different cases.}
Three examples are presented in Figure~\ref{fig:cases} to further inspect the learned weights in the hop encoding. In case 1, a strong clue that matches with the information \textit{``director of''} in question occurs as the mention \textit{``directed by''}, so the weight of mention embedding is relatively high. In case 2, the entity \textit{``World War I''} and \textit{``World War II''} are mentioned with the same context, which means they cannot be distinguished only based on the mention embedding, so more attention is paid to the document embedding which encodes the important fact \textit{``60 million''}. In case 3, no mentions exist in the latest selected document so the hop encoding almost completely depends on the document embedding. We can see that the embedding weights can bring intuitive interpretation about which embedding, or which types of knowledge, is more important for different questions when selecting a hop.

\subsubsection{Probing task for the mention embedding.}
The structured entity relation is represented by the markers around the mentions, as described in Section~\ref{sec:hop_enc}. To explore what the mention embedding learns, we design a special probing task: \textit{distracted hop selection}. That is, the ground-truth hop for bridging questions is shuffled with other hops that have the same mentioned entity but different mention context, and HopRetriever is required to select the right one from these distracting hops for each question. To make the right selection, one should understand more about how each entity is mentioned, but not the entity itself. The summary of this task is shown in Table~\ref{tab:mention}. The experiment result shows that although the distracting hops are not used as negative samples for training, the HopRetriever can retrieve ground-truth hops just based on learned mention embedding at high accuracy (96.42\%), indicating that the mention embedding does learn the implicit relation between entities, but not the entities themselves.

\begin{table}[h]
    \centering
    \footnotesize
    \begin{tabular}{l|c}
        \hline
        Total questions in development set              & 7405    \\ \hline
        Number of the bridging questions                & 5918    \\ \hline
        Average number of distracting hops per question & 52.20   \\ \hline
        \textbf{Accuracy based on mention embedding}    & 96.42\% \\
        \hline
    \end{tabular}
    \caption{Summary of the mention embedding probing task.}
    \label{tab:mention}
\end{table}

\subsubsection{Ablation study.}
As shown in Table~\ref{tab:abl_res}, ablation experiments are conducted to corroborate the effectiveness of HopRetriever. In experiment 1, the structured knowledge in hops is removed (\ie set the weight of mention embedding $w_m$ to 0 in Equation~\ref{eq:fusion}), the performance dropped significantly, which stresses the importance of structured knowledge in Wikipedia for multi-hop evidence retrieval. The performance also degraded in experiment 2 in which the weighting for the structured and unstructured knowledge in hops is disabled  (\ie set $w_m=w_u=1$ in Equation~\ref{eq:fusion}), demonstrating that the fusion function improves the performance while providing interpretations. The auxiliary supporting sentence prediction task is removed in the experiment 3. The result shows that the auxiliary task has no side-effect on the primary hop retrieval task. Additionally, the sentence representations are obtained by the sentence markers contained in the latest retrieved document which has been encoded already at the previous step. So the auxiliary task does not require much additional computation.

\section{Conclusion}
In this paper, we propose the HopRetriever to collect reasoning evidence over Wikipedia for multi-hop question answering. Both the structured knowledge indicated by hyperlinks and the unstructured knowledge presented as introductory documents in Wikipedia, are involved and leveraged together in HopRetriever to help the evidence collection. The experiment on the HotpotQA dataset shows that the performance of HopRetriever improved observably as a result of combining the structured knowledge with unstructured knowledge, and outperforms all the published models on the leaderboard. Moreover, by inspecting the proportion of the two kinds of knowledge in hops, which kind of knowledge leads the retrieving of each evidence piece can be observed directly, which also provides extra intuitive interpretations for the selection of each evidence. 
\bibliography{op_retriever}

\end{document}